\documentclass[runningheads]{llncs}
\usepackage[T1]{fontenc}
\usepackage{graphicx}
\usepackage{amsfonts}
\usepackage{amsmath}
\usepackage{orcidlink}
\usepackage{cleveref}
\usepackage[sort]{cite}
\usepackage{siunitx}
\usepackage{setspace}
\usepackage{nicefrac}
\usepackage{tikz}

\usetikzlibrary{decorations.text}
\usetikzlibrary{decorations.markings}
\usetikzlibrary{calc}
\usetikzlibrary{shapes}
\usetikzlibrary{positioning}
\usetikzlibrary{arrows}
\usetikzlibrary{shapes.arrows}
\usetikzlibrary{arrows.meta}
\usetikzlibrary{fit}
\usetikzlibrary{shapes.geometric}
\usetikzlibrary{backgrounds}
\newlength{\minimumsizeaicon}
\newlength{\minimumheightaicon}
\newlength{\minimumsizeinterconnection}
\newlength{\minimumheightinterconnection}
\newlength{\minimumsizegoal}
\newlength{\roundoffaicon}

\definecolor{aicRed}{HTML}{e78284}
\definecolor{aicGreen}{HTML}{a6d189}
\definecolor{aicBlue}{HTML}{8caaee}
\definecolor{aicOrange}{HTML}{e5c890}
\definecolor{aicGray}{HTML}{dce0e8}
\definecolor{pegGray}{HTML}{a1aeb1}
\definecolor{beadRed}   {RGB}{204,  30,  30}
\definecolor{beadBlue}  {RGB}{ 30,  80, 200}
\definecolor{beadYellow}{RGB}{220, 180,   0}

\tikzset{
	aicon/state/.style={
		rectangle, fill=aicBlue, rounded corners=\roundoffaicon,
		minimum width=\minimumsizeaicon, minimum height=\minimumheightaicon,
		inner sep=0pt, text=white, align=center,
		font=\linespread{0.9}\selectfont
	},
	aicon/active/.style={
		rectangle, rounded corners=\roundoffaicon, fill=aicGreen,
		minimum width=\minimumsizeinterconnection,
		minimum height=\minimumheightinterconnection,
		inner sep=0pt, text=white, align=center,
		font=\linespread{0.9}\selectfont
	},
	aicon/sensor/.style={
		rectangle, rounded corners=\roundoffaicon, fill=aicOrange,
		minimum width=\minimumsizeaicon, minimum height=\minimumheightaicon,
		inner sep=0pt, text=white, align=center,font=\linespread{0.9}\selectfont
	},
	aicon/goal/.style={
		diamond, fill=aicRed, minimum size=\minimumsizegoal,
		inner sep=0pt, text=white
	},
	aicon/link/.style={draw=aicGray, line width=2.5pt},
	aicon/gradient/.style={draw=aicRed, line width=1.8pt, ->, >=latex}
}

\newlength{\tpegsp}     
\newlength{\tpegw}      
\newlength{\tunith}     
\newlength{\tbaseth}    
\newlength{\tmargin}    
\newlength{\tbeadsz}    
\newlength{\tbeadgap}   
\newlength{\tbeadround} 

\newlength{\arrowfromx}
\newlength{\arrowfromy}
\newlength{\arrowtox}
\newlength{\arrowtoy}

\newcommand{\drawTOLscale}[1]{%
  \pgfmathsetlength{\tpegsp}    {#1 * 0.9cm}%
  \pgfmathsetlength{\tpegw}     {#1 * 0.1cm}%
  \pgfmathsetlength{\tunith}    {#1 * 0.45cm}%
  \pgfmathsetlength{\tbaseth}   {#1 * 0.25cm}%
  \pgfmathsetlength{\tmargin}   {#1 * 0.3cm}%
  \pgfmathsetlength{\tbeadsz}   {#1 * 0.38cm}%
  \pgfmathsetlength{\tbeadgap}  {#1 * 0.02cm}%
  \pgfmathsetlength{\tbeadround}{#1 * 0.17cm}%
}

\newcommand{\drawTOLbase}{%
  \fill[pegGray]
    (-\tmargin, 0)
    rectangle ({2*\tpegsp + \tmargin}, -\tbaseth);
  \foreach \i/\h in {0/1, 1/2, 2/3}{
    \fill[pegGray]
      ({\i*\tpegsp - 0.5*\tpegw}, 0)
      rectangle ++(\tpegw, {\h*\tunith});
  }%
}

\newcommand{\drawTOLbeads}[6]{%
  \foreach \col/\px/\py in
    {{#1}/0/0, {#2}/1/0, {#3}/1/1, {#4}/2/0, {#5}/2/1, {#6}/2/2}{%
    \def\tmpcol{\col}%
    \def\tmpnone{none}%
    \ifx\tmpcol\tmpnone
    \else
      \filldraw[fill=\col, draw=\col, rounded corners=\tbeadround]
        ({\px*\tpegsp - 0.5*\tbeadsz},
         {\py*\tbeadsz + \py*\tbeadgap})
        rectangle
        ++({\tbeadsz},{\tbeadsz});
    \fi
  }%
}

\begin{document}
\title{No Plan, Yet Human:\\A Reactive Robotics Model Predicts Human Planning Failures on a Clinical Task}
\titlerunning{No Plan, Yet Human}
%
\author{Michael Migacev\inst{*,1, 2}\orcidlink{0009-0000-7865-9022} \and Vito Mengers\inst{*,1,3}\orcidlink{0000-0001-8341-0299} \and\\
Antonia Köngeter\inst{1}\orcidlink{0009-0003-3700-3993} \and
Oliver Brock\inst{1,3,4}\orcidlink{0000-0002-3719-7754}}
\authorrunning{M. Migacev, V. Mengers, et al.}
%
\institute{Robotics and Biology Laboratory, Technische Universität Berlin, Germany \and 
Institute of Computer Science, Freie Universität Berlin, Germany \and
	Science of Intelligence, Research Cluster of Excellence, Berlin, Germany
	\and Robotics Institute Germany \\$^*$ Equal Contribution\\\url{https://www.tu.berlin/robotics/papers/tower-of-london}}
\maketitle              
\begin{abstract}
Understanding why some sequential planning problems are harder than others requires models that go beyond average performance. They should capture the specific pattern of which problems are hard, and ideally fail in the same way people do when planning capacity is reduced. We apply AICON, a reactive gradient-descent framework developed for robotic manipulation, to the \textit{Tower of London} test, a cognitive test used to assess planning in Parkinson's disease, mild cognitive impairment, and stroke. Without any lookahead planning or knowledge of human cognition, AICON reproduces the fine-grained human difficulty ordering across 24 problems better than structural task parameters and generalizes to held-out problems in a leave-two-out evaluation. Crucially, AICON outperforms a planning baseline for groups with reduced planning capacity while the planning baseline better captures healthy controls. This dissociation was predicted by the original AICON paper, which noted that the model's failure modes resemble those of Parkinson's patients who struggle with goal hierarchies but not move counts. This suggests that as planning capacity is reduced, human behavior shifts toward the reactive mode AICON models. The finding extends a broader pattern: AICON, originally built for robotics, now captures aspects of biological behavior across perception, eye movements, and sequential planning, suggesting its core abstraction reflects something real about how biological systems are organized.
\end{abstract}

\section{Introduction}

When humans solve multi-step problems, they select sequences of actions toward a goal, a capacity typically assumed to require planning: predicting future states to choose actions in the right order~\cite{mattar2022planning}. Yet behavioral evidence suggests that much of what looks like planning may instead emerge from reactive strategies that use the immediate structure of the task directly, without any lookahead~\cite{kirsh1994distinguishing,ballard1997deictic,balleine1998goal}. This tension is hard to resolve by studying healthy people, where planning and reactive strategies are both available and likely intertwined. But people whose planning capacity is reduced by conditions such as Parkinson's disease or stroke offer a different angle: their characteristic failures reveal the underlying system in ways healthy behavior does not. A model that merely fits average difficulty tells us little. One that captures the specific pattern of which problems are hard---and fails in the same way specific people do---provides strong evidence of mechanistic alignment rather than mere data fitting~\cite{kaplan2011explanatory,levenstein2023role}.

The \textit{Tower of London} test (ToL) is a cognitive test that makes this concrete~\cite{shallice1982specific,kaller2012assessing}. Participants rearrange colored beads across three pegs to match a target configuration in as few moves as possible (\Cref{fig:aicon_tol}). People with Parkinson's disease (PD), mild cognitive impairment (MCI), or stroke fail on it in structured, reproducible ways that differ across conditions. These patterns are not fully explained by structural task parameters such as the minimum number of moves required~\cite{kostering2015assessment,mckinlay2008planning}. A mechanistic model capturing this difficulty ordering would help distinguish reactive from planning-based accounts of sequential behavior, and could have practical impact: informing better diagnostic tools and, eventually, targeted interventions.

We pursue this with \textit{Active InterCONnect} (AICON)~\cite{mengers2025noplan}, a framework developed for robotic manipulation that solves sequential tasks without any lookahead. Rather than simulating the future, AICON exploits the current world state through gradient descent, reacting to what is immediately available. We choose AICON for three reasons. First, it was built to solve manipulation problems, not to model human cognition, so any alignment with human failure profiles cannot be a modeling artifact. Second, AICON has already shown surprising biological alignment across perception~\cite{battaje2024information} and scanpath behavior~\cite{mengers2025scanpath}, suggesting its core abstraction captures something real about biological information processing. Third, and most directly: the original AICON paper independently noted that the model's failure modes---struggling with conflicting subgoal hierarchies---resemble the characteristic planning profile of Parkinson's patients~\cite{mengers2025noplan}. That prospective observation is what motivated this study.

We apply AICON to the ToL for the first time, comparing its predicted difficulty ordering against behavioral data from healthy and planning-impaired humans. We show that AICON reproduces the fine-grained human difficulty ordering better than structural parameters. Its failure profile also matches people with impaired planning better than a planning baseline, particularly confirming the prospective Parkinson's disease prediction. Strikingly, the planning baseline better captures healthy controls, revealing a theoretically predicted dissociation, not a modeling accident. We discuss what this implies for how much planning humans actually engage in, for the mechanistic basis of planning deficits in these groups, and for what failure modes tell us about sequential behavior. 

\section{Background \& Related Work}
\subsubsection{The Tower of London Test (ToL)}
is a cognitive test widely used to assess planning deficits in people with Parkinson's disease, mild cognitive impairment, and stroke because these groups exhibit structured, reproducible patterns of failure~\cite{shallice1982specific,kaller2012assessing,kostering2015assessment}. Problem difficulty is commonly characterized by structural parameters such as the minimum number of moves, search depth (how far ahead one must plan before making progress), and the number of suboptimal intermediate states~\cite{kaller2004impact}. Although these parameters explain aggregate performance, they do not specify the cognitive process that generates it. Problems with identical structural parameters can nevertheless differ substantially in human difficulty ordering, making this fine-grained ordering a more demanding criterion for evaluating computational models.

Computational accounts of the ToL have largely assumed explicit planning. ACT-R models explain planning latencies and eye-movement patterns through spreading activation within a cognitive architecture~\cite{albrecht2014spatial}. More recently, Zhang et al.~\cite{zhang2023comparing} compared a range of AI planning algorithms against human behavior, showing that adaptive lookahead planners best account for healthy participants, while classical planners such as BFS perform poorly even when participants are instructed to plan before acting. A follow-up study further showed that planning depth varies systematically across both problems and individuals, with task structure accounting for most of the observed variance~\cite{zhang2025modeling}. Together, these studies suggest that healthy participants engage in adaptive lookahead planning whose depth depends on problem difficulty.

Our work differs in two ways. First, we evaluate a reactive model with no explicit lookahead, asking whether it nonetheless reproduces the human difficulty ordering across problems. Second, we extend the evaluation beyond healthy participants to groups with known planning deficits, which prior computational modeling of the ToL has largely overlooked.
\subsubsection{Active InterCONnect (AICON)}
is a modeling framework originally developed for real-world robotic manipulation~\cite{mengers2025noplan}. Rather than planning ahead by predicting future states, AICON encodes the regularities of a task environment within a network of interconnected components and uses gradient descent to select actions based on the current world state. At each step, AICON computes gradients from a goal cost function through the network and selects the steepest available gradient, implicitly identifying the most relevant subgoal given the current configuration, without any explicit planning.

AICON has been shown to solve long-horizon sequential tasks, including the classic planning domain of Blocks World~\cite{slaney2001blocks}, without any planning~\cite{mengers2025noplan}. In Blocks World, AICON's gradient-based action selection identifies necessary subgoals and solves problems requiring many steps near-optimally, while failing in a characteristic way on problems that require anticipating conflicting subgoals---a failure mode that mirrors human suboptimalities on tower-based tasks~\cite{mengers2025noplan,kaller2004impact,mckinlay2008planning}. Beyond planning tasks, AICON has been applied to real-world robotic perception~\cite{martinmartin2022coupled} and drawer manipulation under uncertainty, where its feedback-driven adaptation outperforms planning-based approaches~\cite{mengers2025noplan}.

Structurally, AICON shares properties with models of biological information processing, and has previously been used to model human visual processing~\cite{battaje2024information}, scanpath behavior~\cite{mengers2025scanpath}, and collective opinion dynamics~\cite{mengers2024leveraging}. Having established AICON as a productive abstraction for biological behavior at the sensory-perceptual level, here we ask whether it extends to sequential planning behavior on a clinical task and what this reveals about the mechanisms underlying human planning and its characteristic failures.

\section{Methods}

We build an AICON model that encodes the rules of the ToL as a network of interconnected estimators and evaluate it against human behavioral data from four groups across 24 problems. We describe the model, the evaluation procedure, and the baselines in turn.

\subsection{An AICON Model for Tower of London Tasks}\label{sec:model}

A typical Tower of London problem consists of three pegs of ascending height, with three colored beads arranged across them. We represent the board state as a $6 \times 3$ matrix $\mathbf{x}$, where the six rows correspond to the six possible positions across the three pegs (one position for the first peg, two for the second peg etc.) and the three columns encode bead color in a one-hot scheme.

\textbf{Encoding the Rules of Tower of London}\quad
The AICON model for the ToL consists of three recursive estimators and three active interconnections, shown in \Cref{fig:aicon_tol} (bottom left). The \textit{State} estimator $\mathbf{x} \in [0,1]^{6 \times 3}$ is the only estimator directly connected to the world: it receives the current board configuration as a sensor input and serves as the sole interface between the physical problem state and the rest of the network. The remaining two recursive estimators track the quantities required to determine which moves are legal at any given state: the \textit{Movable} estimator $\mathbf{m} \in [0,1]^{6}$ tracks which positions contain a bead that can be picked up, and the \textit{Free} estimator $\mathbf{f} \in [0,1]^{6}$ tracks which positions can legally receive a bead. While the board state sensor provides a discrete one-hot encoding of bead positions, all recursive estimators maintain soft values in $[0,1]$, treating occupancy and legality as probabilities rather than hard assignments to ensure the network remains differentiable throughout and gradients can propagate to drive action selection.

A position contributes to $\mathbf{m}$ if and only if it is occupied and has no bead above it. The \textit{Exposed Beads} active interconnection computes this from the current state $\mathbf{x}$ by combining the occupancy vector with a static structural mask encoding the vertical relationships between positions on each peg. Similarly, a position contributes to $\mathbf{f}$ if and only if it is empty and either is at the bottom of its peg or has an occupied position directly below it. The \textit{Supported Empty Fields} active interconnection computes this from $\mathbf{x}$ using the complementary occupancy and the transpose of the structural mask.

The action $\mathbf{a} \in [0,1]^{6 \times 6}$ encodes a move as a from-position to-position matrix. The \textit{Legal Move Constraint} couples $\mathbf{a}$, $\mathbf{m}$, and $\mathbf{f}$ to constrain the action to feasible moves and propagate the resulting change back into the state $\mathbf{x}$.

\begin{figure}[t]
	\centering
	\newlength{\tolnodedisty}
	\newlength{\tolnodedistx}
	\setlength{\tolnodedisty}{1cm}
	\setlength{\tolnodedistx}{2.15cm}
	\pgfdeclarelayer{bg}
	\pgfsetlayers{bg,main}
	\small
	\begin{tikzpicture}[on grid,
		node distance=\tolnodedisty and \tolnodedistx]
		
		\node[aicon/sensor] (boardstate) {Board State\\Sensor};
		\node[aicon/sensor, below=\tolnodedisty of boardstate] (action) {Action $\mathbf{a}$\\{\scriptsize$6{\times}6$}};
		
		\node[aicon/state, right=of boardstate, yshift=0.5\tolnodedisty] (movable) {Movable $\mathbf{m}$\\{\scriptsize$6{\times}1$}};
		\node[aicon/active, below=\tolnodedisty of movable] (legalmove) {Legal Move\\Constraint};
		\node[aicon/state, below=\tolnodedisty of legalmove] (free) {Free $\mathbf{f}$\\{\scriptsize$6{\times}1$}};
		
		\node[aicon/active, right=of movable] (exposedbeads) {Exposed\\Beads};
		\node[aicon/state, below=\tolnodedisty of exposedbeads] (state) {State $\mathbf{x}$\\{\scriptsize$6{\times}3$}};
		\node[aicon/active, below=\tolnodedisty of state] (supportedempty) {Supported\\Empty Fields};
		
		\node[aicon/goal, right=0.625\tolnodedistx of state] (g) {$g$};
		
		\node[below=1.15\tolnodedisty of action,align=left,anchor=west,xshift=-0.8cm] {\textbf{AICON Model}};
		\node[above=2.75\tolnodedisty of boardstate,align=left,anchor=west,xshift=-0.8cm] {\textbf{Example ToL Problem}};
		\node[right=\tolnodedistx of exposedbeads,align=left,anchor=center,xshift=-0.1cm,yshift=0.0cm,rotate=90] {\textbf{Action Gradients}};
		
		\begin{pgfonlayer}{bg}
			\draw[aicon/link] (boardstate.center) -- ([xshift=-0.5\tolnodedistx,yshift=0.5\tolnodedisty]state.center)
			-- (state.center);
			
			\draw[aicon/link] (action.center) -- ([xshift=0.5\tolnodedistx]action.center) |- (legalmove.center);
			
			\draw[aicon/link] (state.center) -- (exposedbeads.center);
			
			\draw[aicon/link] (state.center) -- (supportedempty.center);
			
			\draw[aicon/link] (exposedbeads.center) -- (movable.center);
			
			\draw[aicon/link] (supportedempty.center) -- (free.center);
			
			\draw[aicon/link] (movable.center) -- (legalmove.center);
			
			\draw[aicon/link] (free.center) -- (legalmove.center);
			
			\draw[aicon/link] (legalmove.center) -- (state.center);
			
			\draw[aicon/link] (state.center) -- (g.center);
		\end{pgfonlayer}
		
		\begin{scope}[shift={(0.5cm, 1.5cm)}, node distance=0.45cm and 1cm]
			\drawTOLscale{0.7}
			\drawTOLbase
			\drawTOLbeads{beadRed}{none}{none}{beadYellow}{beadBlue}{none}
		\end{scope}
		\begin{scope}[shift={(3.8cm, 1.5cm)}, node distance=0.45cm and 1cm]
			\drawTOLscale{0.7}
			\drawTOLbase
			\drawTOLbeads{beadBlue}{beadRed}{beadYellow}{none}{none}{none}
		\end{scope}
		\node[above=1.25\tolnodedisty of movable,xshift=-2.5cm] {Start};
		\node[above=1.25\tolnodedisty of movable,xshift=0.8cm] {Goal};
		
		\begin{scope}[shift={(7cm, 2.5cm)}, node distance=0.45cm and 0.9cm]
			\setlength{\minimumsizeaicon}{0.7cm}
			\setlength{\minimumheightaicon}{0.25cm}
			\setlength{\minimumsizeinterconnection}{0.9cm}
			\setlength{\minimumheightinterconnection}{0.3cm}
			\setlength{\minimumsizegoal}{10pt}
			\setlength{\roundoffaicon}{1pt}
			\node[aicon/sensor] (Abs) {};
			\node[aicon/sensor, below= of Abs] (Aa) {};
			\node[aicon/state, right=of Abs, yshift=0.275cm] (Am) {};
			\node[aicon/active, below=of Am] (Alm) {};
			\node[aicon/state, below=of Alm] (Af) {};
			\node[aicon/active, right=of Am] (Aeb) {};
			\node[aicon/state, below=of Aeb] (Ax) {};
			\node[aicon/active, below=of Ax] (Ase) {};
			\node[aicon/goal, right=0.72cm of Ax] (Ag) {};
			\begin{pgfonlayer}{bg}
				\draw[aicon/link] ([yshift=0.05cm]Abs.center) -- ([xshift=-0.4cm,yshift=0.2cm]Ax.center) -- (Ax.center);
				\draw[aicon/link] (Aa.center) -- ([xshift=0.45cm]Aa.center) |- (Alm.center);
				\draw[aicon/link] (Ax.center) -- (Aeb.center);
				\draw[aicon/link] (Ax.center) -- (Ase.center);
				\draw[aicon/link] (Aeb.center) -- (Am.center);
				\draw[aicon/link] (Ase.center) -- (Af.center);
				\draw[aicon/link] (Am.center) -- (Alm.center);
				\draw[aicon/link] (Af.center) -- (Alm.center);
				\draw[aicon/link] (Alm.center) -- (Ax.center);
				\draw[aicon/link] (Ax.center) -- (Ag.center);
			\end{pgfonlayer}
			\draw[aicon/gradient] (Ag.center) -- (Ax.center);
			\draw[aicon/gradient] (Ax.center) -- (Alm.center);
			\draw[aicon/gradient] (Alm.center) -- (Aa.center);
			\node[below=of Ase] {$\nabla_{\mathbf{a}}^{\mathrm{p}_1}\, g =
				\frac{\partial g}{\partial \mathbf{x}}
				\frac{\partial \mathbf{x}}{\partial \mathbf{a}}$};
		\end{scope}
		
		\begin{scope}[shift={(10cm, 1.85cm)}, node distance=0.45cm and 1cm]
			\drawTOLscale{0.5}
			\drawTOLbase
			\drawTOLbeads{beadRed}{none}{none}{beadYellow}{beadBlue}{none}
			\pgfmathsetlength{\arrowtox}{1 * \tpegsp}
			\pgfmathsetlength{\arrowtoy}{\tbeadgap + 0.1 * \tbeadsz}
			\pgfmathsetlength{\arrowfromx}{0.1 * \tbeadsz}
			\pgfmathsetlength{\arrowfromy}{1.1 * \tbeadsz}
			
			\draw[-stealth, draw=aicRed, line width=1.0pt, bend right=-60]
			(\the\arrowfromx, \the\arrowfromy) to (\the\arrowtox, \the\arrowtoy);
		\end{scope}
		
		\begin{scope}[shift={(7cm,0.65cm)}, node distance=0.45cm and 0.9cm]
			\setlength{\minimumsizeaicon}{0.7cm}
			\setlength{\minimumheightaicon}{0.25cm}
			\setlength{\minimumsizeinterconnection}{0.9cm}
			\setlength{\minimumheightinterconnection}{0.3cm}
			\setlength{\minimumsizegoal}{10pt}
			\setlength{\roundoffaicon}{1pt}
			\node[aicon/sensor] (Abs) {};
			\node[aicon/sensor, below= of Abs] (Aa) {};
			\node[aicon/state, right=of Abs, yshift=0.275cm] (Am) {};
			\node[aicon/active, below=of Am] (Alm) {};
			\node[aicon/state, below=of Alm] (Af) {};
			\node[aicon/active, right=of Am] (Aeb) {};
			\node[aicon/state, below=of Aeb] (Ax) {};
			\node[aicon/active, below=of Ax] (Ase) {};
			\node[aicon/goal, right=0.72cm of Ax] (Ag) {};
			\begin{pgfonlayer}{bg}
				\draw[aicon/link] ([yshift=0.05cm]Abs.center) -- ([xshift=-0.4cm,yshift=0.2cm]Ax.center) -- (Ax.center);
				\draw[aicon/link] (Aa.center) -- ([xshift=0.45cm]Aa.center) |- (Alm.center);
				\draw[aicon/link] (Ax.center) -- (Aeb.center);
				\draw[aicon/link] (Ax.center) -- (Ase.center);
				\draw[aicon/link] (Aeb.center) -- (Am.center);
				\draw[aicon/link] (Ase.center) -- (Af.center);
				\draw[aicon/link] (Am.center) -- (Alm.center);
				\draw[aicon/link] (Af.center) -- (Alm.center);
				\draw[aicon/link] (Alm.center) -- (Ax.center);
				\draw[aicon/link] (Ax.center) -- (Ag.center);
			\end{pgfonlayer}
			\draw[aicon/gradient] (Ag.center) -- ([yshift=-0.1cm,xshift=0.1cm]Ax.center);
			\draw[aicon/gradient] ([yshift=-0.1cm,xshift=0.1cm]Ax.center) -- ([yshift=-0.1cm]Alm.center);
			\draw[aicon/gradient] ([yshift=-0.1cm]Alm.center) -- ([yshift=-0.1cm]Af.center);
			\draw[aicon/gradient] ([yshift=-0.1cm]Af.center) -- ([xshift=-0.2cm,yshift=-0.1cm]Ase.center);
			\draw[aicon/gradient] ([xshift=-0.2cm,yshift=-0.1cm]Ase.center) -- ([yshift=0.1cm,xshift=-0.2cm]Ax.center);
			\draw[aicon/gradient] ([yshift=0.1cm,xshift=-0.2cm]Ax.center) -- ([yshift=0.1cm,xshift=-0.2cm]Alm.center);
			\draw[aicon/gradient] ([yshift=0.1cm,xshift=-0.2cm]Alm.center) -- (Aa.center);
			\node[below=of Ase] {$	\nabla_{\mathbf{a}}^{\mathrm{p}_2}\, g =
				\frac{\partial g}{\partial \mathbf{x}}
				\frac{\partial \mathbf{x}}{\partial \mathbf{f}}
				\frac{\partial \mathbf{f}}{\partial \mathbf{x}}
				\frac{\partial \mathbf{x}}{\partial \mathbf{a}}$};
		\end{scope}
		\begin{scope}[shift={(10cm, 0cm)}, node distance=0.45cm and 1cm]
			\drawTOLscale{0.5}
			\drawTOLbase
			\drawTOLbeads{beadRed}{none}{none}{beadYellow}{beadBlue}{none}
			\pgfmathsetlength{\arrowtox}{1 * \tpegsp}
			\pgfmathsetlength{\arrowtoy}{\tbeadgap + 0.1 * \tbeadsz}
			\pgfmathsetlength{\arrowfromx}{0.1 * \tbeadsz}
			\pgfmathsetlength{\arrowfromy}{1.1 * \tbeadsz}
			
			\draw[-stealth, draw=aicRed, line width=1.0pt, bend right=-60]
			(\the\arrowfromx, \the\arrowfromy) to (\the\arrowtox, \the\arrowtoy);
						\pgfmathsetlength{\arrowfromx}{2 * \tpegsp - 0.1 * \tbeadsz}
			\pgfmathsetlength{\arrowfromy}{\tbeadsz + \tbeadgap + 1.1 * \tbeadsz}
			
			\draw[-stealth, draw=aicRed, line width=1.0pt, bend right=50]
			(\the\arrowfromx, \the\arrowfromy) to (\the\arrowtox, \the\arrowtoy);
		\end{scope}
		
		\begin{scope}[shift={(7cm,-1.2cm)}, node distance=0.45cm and 0.9cm]
			\setlength{\minimumsizeaicon}{0.7cm}
			\setlength{\minimumheightaicon}{0.25cm}
			\setlength{\minimumsizeinterconnection}{0.9cm}
			\setlength{\minimumheightinterconnection}{0.3cm}
			\setlength{\minimumsizegoal}{10pt}
			\setlength{\roundoffaicon}{1pt}
			\node[aicon/sensor] (Abs) {};
			\node[aicon/sensor, below= of Abs] (Aa) {};
			\node[aicon/state, right=of Abs, yshift=0.275cm] (Am) {};
			\node[aicon/active, below=of Am] (Alm) {};
			\node[aicon/state, below=of Alm] (Af) {};
			\node[aicon/active, right=of Am] (Aeb) {};
			\node[aicon/state, below=of Aeb] (Ax) {};
			\node[aicon/active, below=of Ax] (Ase) {};
			\node[aicon/goal, right=0.72cm of Ax] (Ag) {};
			\begin{pgfonlayer}{bg}
				\draw[aicon/link] ([yshift=0.05cm]Abs.center) -- ([xshift=-0.4cm,yshift=0.2cm]Ax.center) -- (Ax.center);
				\draw[aicon/link] (Aa.center) -- ([xshift=0.45cm]Aa.center) |- (Alm.center);
				\draw[aicon/link] (Ax.center) -- (Aeb.center);
				\draw[aicon/link] (Ax.center) -- (Ase.center);
				\draw[aicon/link] (Aeb.center) -- (Am.center);
				\draw[aicon/link] (Ase.center) -- (Af.center);
				\draw[aicon/link] (Am.center) -- (Alm.center);
				\draw[aicon/link] (Af.center) -- (Alm.center);
				\draw[aicon/link] (Alm.center) -- (Ax.center);
				\draw[aicon/link] (Ax.center) -- (Ag.center);
			\end{pgfonlayer}
			\draw[aicon/gradient] (Ag.center) -- ([yshift=0.1cm,xshift=0.1cm]Ax.center);
			\draw[aicon/gradient] ([yshift=0.1cm,xshift=0.1cm]Ax.center) -- ([yshift=0.1cm]Alm.center);
			\draw[aicon/gradient] ([yshift=0.1cm]Alm.center) -- ([yshift=0.1cm]Am.center);
			\draw[aicon/gradient] ([yshift=0.1cm]Am.center) -- ([xshift=-0.2cm,yshift=0.1cm]Aeb.center);
			\draw[aicon/gradient] ([xshift=-0.2cm,yshift=0.1cm]Aeb.center) -- ([yshift=-0.1cm,xshift=-0.2cm]Ax.center);
			\draw[aicon/gradient] ([yshift=-0.1cm,xshift=-0.2cm]Ax.center) -- ([yshift=-0.1cm,xshift=-0.2cm]Alm.center);
			\draw[aicon/gradient] ([yshift=-0.1cm,xshift=-0.2cm]Alm.center) -- (Aa.center);
			\node[below=of Ase] {$\nabla_{\mathbf{a}}^{\mathrm{p}_3}\, g =
				\frac{\partial g}{\partial \mathbf{x}}
				\frac{\partial \mathbf{x}}{\partial \mathbf{m}}
				\frac{\partial \mathbf{m}}{\partial \mathbf{x}}
				\frac{\partial \mathbf{x}}{\partial \mathbf{a}}$};
		\end{scope}
				\begin{scope}[shift={(10cm, -1.85cm)}, node distance=0.45cm and 1cm]
			\drawTOLscale{0.5}
			\drawTOLbase
			\drawTOLbeads{beadRed}{none}{none}{beadYellow}{beadBlue}{none}
			\pgfmathsetlength{\arrowtox}{1 * \tpegsp}
			\pgfmathsetlength{\arrowtoy}{\tbeadgap + 0.1 * \tbeadsz}
			\pgfmathsetlength{\arrowfromx}{2 * \tpegsp - 0.1 * \tbeadsz}
			\pgfmathsetlength{\arrowfromy}{\tbeadsz + \tbeadgap + 1.1 * \tbeadsz}
			
			\draw[-stealth, draw=aicRed, line width=1.0pt, bend right=50]
			(\the\arrowfromx, \the\arrowfromy) to (\the\arrowtox, \the\arrowtoy);
		\end{scope}
		
	\end{tikzpicture}
	\caption{AICON model (bottom left) for the \textit{Tower of London} test (top left). In the model, board state and action are sensors encoding the current world state. Movable and Free are recursive estimators tracking which beads can be picked up and which fields can receive a bead respectively. Exposed Beads and Supported Empty Fields are active interconnections computing these quantities from the current state. Legal Move Constraint is an active interconnection coupling action, Movable, and Free to update the state. The goal $g$ drives action selection to the goal state by descending the steepest gradient through the network. On the right, we show the potential gradients for the example, corresponding to ``\textit{move red into goal position}'' (top), ``\textit{ensure there is a bead supporting yellow's goal position and/or that blue's goal position is empty}'' (middle), and ``\textit{ensure yellow is not blocked from above}'' (bottom).}
	\label{fig:aicon_tol}
\end{figure}

\textbf{Goal Function and its Gradients}\quad
The goal $g$ is defined as a differentiable cost function over the state estimator $\mathbf{x}$, penalizing deviation from the target configuration. To ensure gradients prioritize resolving the lowest occupied peg level first, we utilize ``smart'' goals~\cite{mengers2025noplan}, making the target configuration depend on the current state rather than specifying a fixed target throughout. To select an action at each step, AICON computes gradients of $g$ with respect to $\mathbf{a}$ along all available paths through the network and selects the steepest~\cite{mengers2025noplan}. The available gradient paths correspond directly to the types of moves the system can consider.

As shown in \Cref{fig:aicon_tol} (right), path $\mathrm{p}_1$ corresponds to directly moving a bead toward the goal configuration. Paths $\mathrm{p}_2$ and $\mathrm{p}_3$ correspond to one level of subgoal: freeing a target field or removing a blocking bead respectively. These three paths can be composed by alternating $\mathbf{m}$ and $\mathbf{f}$ terms to identify actionable subgoals at arbitrary depth, e.g., needing to move a bead to free a field before another bead can be placed. In all cases, the steepest gradient across all available paths is selected at each step~\cite{mengers2025noplan}, identifying the most immediately actionable move without any explicit forward search. The magnitude of gradient chains diminishes as the chain lengthens, producing weaker signals for problems that require longer sequences of preparatory moves---mirroring the effect of search depth on human performance and providing a mechanistic account of AICON's difficulty ordering.

\subsection{Comparing Model and Human Behavior}
\indent\indent
\textbf{Human Behavioral Data}\quad
We evaluate AICON against human behavioral data from K\"ostering et al.~\cite{kostering2015assessment,kostering2016analyses}, which provides performance on a standardized set of 24 ToL problems across four groups: healthy controls ($n = 155$), and patients with Parkinson's disease (PD) ($n = 51$), mild cognitive impairment (MCI) ($n = 29$), and after a stroke ($n = 60$). The original dataset includes an additional 4 practice problems which we exclude from all analyses. For each participant and problem, the dataset records success and number of additional moves taken beyond the optimal solution. We aggregate these measures per problem and group to obtain a difficulty ordering across the 24 problems, which serves as our behavioral target for model comparison.

\textbf{Evaluation Metrics \& Parameter Identification}\quad
We evaluate model-human alignment using Kendall's $\tau$, a rank correlation measure~\cite{kendall1938new} that captures the degree to which the model reproduces the human difficulty ordering. AICON produces average additional moves per problem as its primary output. We correlate this against two human behavioral measures independently: success rate (proportion of participants solving each problem) and average additional moves beyond optimal (for successful participants). Both human measures are computed per group and per problem, and $\tau$ is reported for each, allowing us to assess whether the model captures both dimensions of human difficulty.

AICON has two free parameters $\boldsymbol{\alpha}$ and $\boldsymbol{\beta}$, which scale the change made to their respective recursive estimators $\mathbf{f}$ and $\mathbf{m}$. To avoid overfitting these parameters to the full problem set, we adopt a leave-two-out generalization procedure: parameters are identified by maximizing Kendall's $\tau$ on 22 of the 24 problems, and the model is then evaluated on the two held-out problems. Due to computational constraints, we sample $N = 120$ randomly selected splits from the $\binom{24}{2} = 276$ possible held-out pairs, using a fixed random seed for reproducibility. For each split, the held-out pair yields a $\tau \in \{-1, 0, 1\}$ depending on whether the model orders the two problems correctly, incorrectly, or assigns them equal difficulty. We report the average $\tau$ across splits as our generalization score, alongside the in-sample $\tau$ on the 22 fitting problems. This procedure tests whether the model's difficulty ordering generalizes to unseen problems rather than merely fitting the problems it was tuned on.

\textbf{Problem Parameters and Planning as Baselines}\quad
We compare AICON against two types of baseline. The first is a classic descriptive parameter: the optimal number of moves required for each problem, which represents the standard account of ToL difficulty and serves as a lower bound.

\begin{figure}[t]
	\centering
	\includegraphics[width=\textwidth]{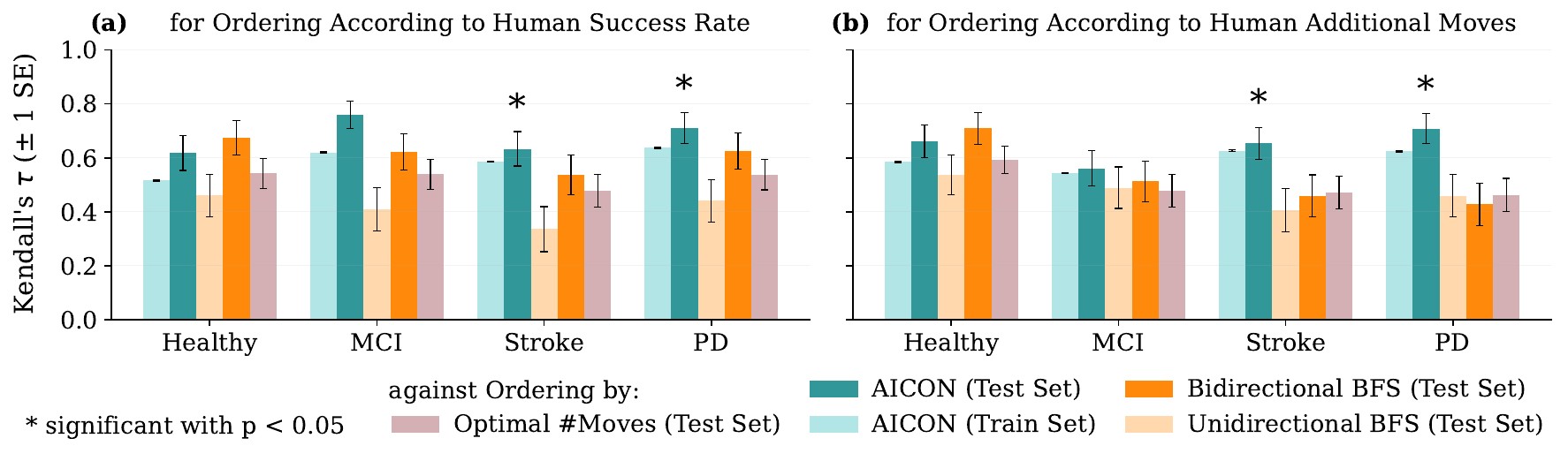}
	\caption{AICON reproduces the fine-grained human difficulty ordering and outperforms planning baselines for groups with reduced planning capacity. Asterisks indicate significance when comparing the AICON test set to all baselines. Significance is assessed via paired permutation tests ($10^6$ permutations) and FDR-corrected $p < 0.05$ (32 comparisons) across all baselines. Kendall's $\tau$ between model-predicted and human difficulty ordering (success rate: \textbf{a}; additional moves: \textbf{b}) is shown for AICON on fitting problems (Train) and held-out problems (Test), alongside bidirectional BFS, unidirectional BFS, and optimal moves as baselines on the Test set. AICON generalizes robustly: Train and Test $\tau$ are consistently close across all groups. Bidirectional BFS outperforms unidirectional BFS consistently, motivating its use as the primary planning comparison. For healthy controls, BFS matches or exceeds AICON, consistent with healthy participants engaging in lookahead planning. For groups with reduced planning capacity, particularly PD, AICON's held-out $\tau$ exceeds both BFS variants, especially on additional moves (\textbf{b}), while optimal moves as a descriptive account ranks lowest throughout.}
	\label{fig:results_total}
\end{figure}

The second is a planning baseline based on breadth-first search (BFS). A unidirectional BFS, searching from either the start state or the goal state, captures little of the human difficulty ordering, as it reflects only one direction of the search process. We therefore also evaluate a bidirectional BFS baseline, which averages the number of nodes visited when searching simultaneously from both the start and goal states until the two searches meet. As shown in \Cref{fig:results_total}, bidirectional BFS captures human difficulty ordering substantially better than its unidirectional counterpart, motivating its use as our primary planning baseline. Both BFS variants are included in the figures for transparency.

\section{Results}

We evaluated AICON's ability to reproduce the human difficulty ordering across 24 ToL problems for four groups (healthy, PD, MCI, stroke), comparing against three baselines: unidirectional BFS, bidirectional BFS (both representing pure planning strategies differing in search direction), and optimal moves (the minimum number of moves required, a descriptive parameter for each problem). For AICON, results are reported both as Kendall's $\tau$ on the 22 fitting problems (train set) and as average $\tau$ on held-out problem pairs across $N=120$ randomly sampled leave-two-out splits (test set). For all baselines, which have no free parameters, only the held-out $\tau$ is reported, keeping comparisons on equal~footing.

\subsection{AICON Captures Human Difficulty Ordering}

\Cref{fig:results_total} shows that AICON achieves consistent positive $\tau$ across all four groups on both the fitting problems (light blue) and held-out problems (dark blue), and that this holds for both behavioral measures. This suggests AICON captures a general difficulty signal that generalizes beyond the problems it was tuned on.

\Cref{fig:results_splits} breaks generalization down by the difficulty of the held-out pair, where difficulty was determined by the average success rate across human groups. When the two held-out problems span difficulty levels (easy vs.\ hard, middle), generalization is near-perfect across all groups and models---any reasonable model can distinguish easy from hard problems. The informative result is in the within-difficulty columns. For easy vs.\ easy pairs (left), AICON maintains positive $\tau$ across all groups. Within the hardest problems (right), generalization drops and becomes more variable for all models, consistent with the mechanistic account that gradient signals weaken for problems requiring long subgoal chains. Significance is reached for PD and stroke on both behavioral measures (paired permutation test, FDR-corrected $p < 0.05$ across 32 comparisons; asterisks in \Cref{fig:results_total}), but not for healthy controls or MCI.

\subsection{AICON Outperforms Baselines for Planning-Impaired Groups}

\begin{figure}[t]
	\centering
	\includegraphics[width=\textwidth]{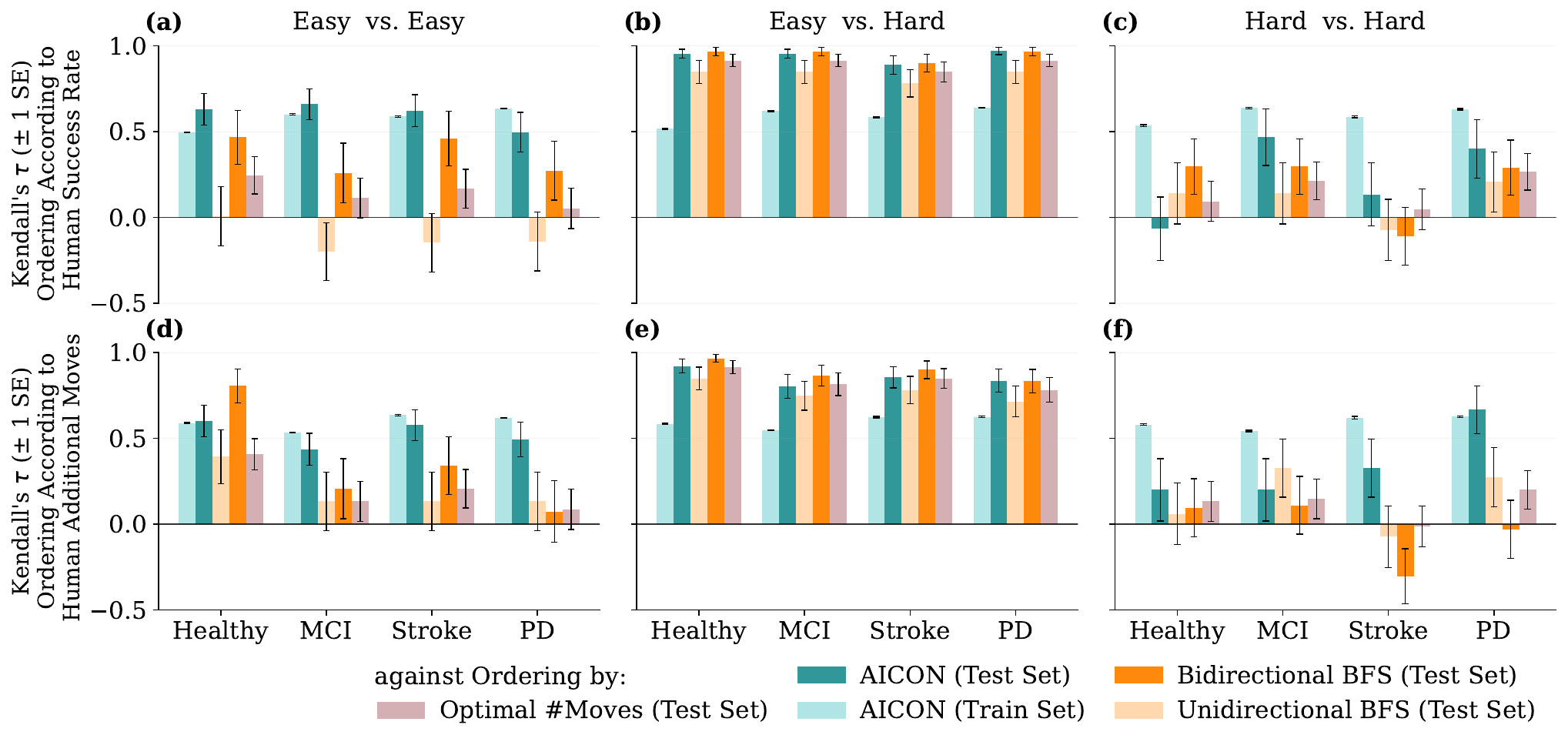}
	\caption{Within-difficulty generalization reveals where AICON's reactive mechanism has an advantage. Held-out $\tau$ is shown separately for easy vs.\ easy (left), easy vs.\ hard (middle), and hard vs.\ hard (right) splits, for success rate (top row) and additional moves (bottom row). Easy vs.\ hard generalization is near-perfect for all models---distinguishing easy from hard problems is trivial. For easy pairs, AICON maintains positive $\tau$ while BFS variants degrade sharply, particularly for groups with reduced planning capacity. Within the hardest problems, generalization drops for all models, but BFS variants degrade more substantially for planning-impaired groups while AICON remains comparatively stable. Due to the smaller number of pairs per split, variance is higher and no individual comparison reaches significance after FDR correction.}
	\label{fig:results_splits}
	
\end{figure}

The key comparison is between AICON and BFS. Returning to \Cref{fig:results_total}, for healthy controls the BFS bars match or exceed AICON on the test set across both measures, suggesting that a planning baseline captures healthy behavior at least as well as a reactive model. For planning-impaired groups the pattern reverses: AICON (dark blue) consistently exceeds bidirectional BFS (dark orange), most clearly for PD on additional moves (\Cref{fig:results_total}\textbf{b}). The advantage holds for MCI and stroke as well, though it is somewhat smaller and more variable across splits.

This dissociation is visible across difficulty splits in \Cref{fig:results_splits}. For easy vs.\ easy pairs (left), BFS degrades sharply for planning-impaired groups while AICON remains positive---particularly on additional moves (bottom). This is the regime where the reactive gradient mechanism has an advantage: BFS assigns similar difficulty to problems that have similarly sized search spaces but different sequential structure, whereas AICON's gradient chains are sensitive to that structure.

The optimal moves baseline performs consistently below both AICON and BFS across all groups, confirming that structural task parameters alone do not account for the fine-grained difficulty ordering.

Taken together, these results suggest that as planning capacity is reduced---whether through Parkinson's disease, mild cognitive impairment, or stroke---human behavior shifts toward the reactive mode that AICON models. The reversal of the AICON-BFS relationship between healthy controls and planning-impaired groups is not a general modeling advantage but a specific, theoretically predicted signature of mechanistic alignment.

\section{Discussion}

The result that AICON best matches Parkinson's patients among the four groups is not a post-hoc observation but the confirmation of a prediction made before this study. The original AICON paper noted that the model struggles with conflicting subgoal hierarchies while staying competent on other problems, much like PD patients~\cite{mengers2025noplan}. That observation motivated the present study, and the data confirm it. A framework developed for robotic manipulation, applied to a clinical planning task with no knowledge of human neuroscience, reproduces the specific failure profile of people with known basal ganglia dysfunction.

\textbf{Planning, Reacting, and What AICON Models}\quad
The dissociation between healthy and clinical groups touches on a longstanding question in cognitive science~\cite{kirsh1994distinguishing,ballard1997deictic}: how much of sequential problem solving depends on explicit lookahead planning, and how much emerges from reactive strategies? Our results suggest the answer depends on who you are asking. For healthy adults, a planning baseline matches their difficulty ordering at least as well as AICON, consistent with healthy participants using meaningful lookahead on the ToL~\cite{waldron2011influence,zhang2023comparing}. For planning-impaired groups---particularly PD and stroke, but also MCI---AICON's reactive mechanism provides a better account, suggesting that as planning capacity is reduced, behavior might shift toward the reactive mode AICON models.

AICON captures one component of this picture: the reactive use of immediately available task structure. What it does not capture is the deliberate planning that healthy participants appear to use. This is not a failure of the model but a feature of its design: AICON was built without lookahead, and the fact that it still achieves positive generalization even for healthy participants suggests that reactive strategies contribute meaningfully to human performance even when planning capacity is intact. The two components are not mutually exclusive---humans likely use both, with the balance shifting as planning capacity changes.

The harder problems---where AICON's generalization is weakest---are exactly those that require anticipating conflicting subgoals. This is where the gradient-based mechanism breaks down, producing weak action signals. That this is also where human performance is most variable is telling. It is also where planning-impaired groups struggle most, consistent with these problems requiring deliberate planning resources that AICON does not model.

\textbf{Mechanism Beats Description}\quad
A recurring theme in computational cognitive modeling is the tension between descriptive and mechanistic accounts~\cite{levenstein2023role}. Descriptive models, such as the task parameters we use as a baseline, identify variables that predict behavior but do not specify the process that generates it. As our results show, they fail to capture the fine-grained problem ordering that distinguishes individual problems within the same parameter range. AICON, by contrast, commits to a specific mechanism: reactive gradient descent over task regularities, constrained by what is available at each step. Our mechanistic model makes the PD prediction structurally, because its failure modes are determined by its architecture rather than by the data it was fit to---a descriptive model can only describe what it sees. This is the sense in which getting the mechanism right matters: not just for better prediction, but for the kind of science it enables~\cite{kaplan2011explanatory}.

\textbf{Limitations}\quad First, AICON models a single component of human problem solving---reactive action selection---and does not currently account for individual differences, learning, or the deliberate planning strategies that healthy participants appear to use. A complete model would need both components and would need to explain how their balance shifts across groups and problem types.

Second, our evaluation relies on a single dataset~\cite{kostering2015assessment} with fixed problem and participant group structure. Generalization across different ToL problem sets, clinical instruments, or groups remains to be established.

Third, the leave-two-out generalization procedure is limited by the binary nature of each held-out evaluation, and the weak generalization within the hardest problems suggests that AICON's difficulty ordering in this regime may not yet be reliable enough for clinical use. Future work connecting AICON's failure modes to neuroimaging or lesion data would also strengthen the mechanistic claim that the PD result reflects basal ganglia-dependent goal hierarchy processing.

More broadly, a validated mechanistic model of ToL performance could inform clinical applications: identifying problems that maximally discriminate between groups, tracking disease progression through shifts in difficulty profiles, or informing the design of targeted interventions. These applications remain a longer-term goal contingent on the model's further validation.

\section{Conclusion}

We have shown that AICON, a modeling framework developed for robotic manipulation, captures the fine-grained human difficulty ordering on the \textit{Tower of London} test better than structural task parameters, and better than a planning baseline for groups with reduced planning capacity. For healthy people, the planning baseline wins---consistent with them using lookahead. For planning-impaired groups, AICON wins---consistent with their behavior being better described by reactive strategies. This reversal was predicted from the model's architecture before the study was conducted~\cite{mengers2025noplan}, not fitted to the data. We take this as evidence that AICON captures something real about the reactive component of human sequential problem solving, one that becomes especially visible when planning capacity is reduced.

This result is part of a broader pattern: AICON, originally built for robotics, has now been shown to capture aspects of biological behavior at the level of visual perception~\cite{battaje2024information}, eye movement behavior~\cite{mengers2025scanpath}, collective opinion dynamics~\cite{mengers2024leveraging}, and now sequential planning. That a single framework spans these levels suggests its core abstraction captures something real about how information processing in biological systems is organized, in ways that go beyond the boundary between robotic and biological intelligence.

\begin{credits}
\subsubsection{\ackname} We thank Dr.\ Christoph Kaller and Dr.\ Lena Schumacher for generously sharing and explaining the behavioral data from their previous studies on the ToL as a diagnostic tool~\cite{kostering2015assessment,kostering2016analyses}. We gratefully acknowledge funding by the Deutsche Forschungsgemeinschaft (DFG, German Research Foundation) under Germany’s Excellence Strategy – EXC 2002/1 “Science of Intelligence” – project number 390523135. This work has been partially supported by the German Federal Ministry of Research, Technology and Space (BMFTR) under the Robotics Institute Germany~(RIG). 

\subsubsection{\discintname}
The authors declare no competing interests.

\subsubsection{Code Availability} for model and analysis: \url{https://github.com/tu-rbo/tol-aicon}

\end{credits}

\bibliographystyle{splncs04}
\bibliography{aicon_towers}

\end{document}